\documentclass{isprs}
\usepackage{subfigure}
\usepackage{setspace}
\usepackage{geometry}
\usepackage{epstopdf}
\usepackage[labelsep=period]{caption}
\usepackage[british]{babel}
\usepackage{booktabs}
\usepackage{url}
\usepackage{comment}
\usepackage{microtype}

\usepackage{algorithm,algorithmic}

\usepackage[T1]{fontenc}
\usepackage{xcolor}

\begin{document}

\title{BrazilDAM: A Benchmark dataset for Tailings Dam Detection}

% KAO: Remove extra spacing
%\author{PAPER ID 1118}
%\author{DOUBLE BLIND}
\author{Edemir Ferreira\textsuperscript{1}, Matheus Brito\textsuperscript{1}, Remis Balaniuk\textsuperscript{2}, M\'ario S. Alvim\textsuperscript{1}, Jefersson A. dos Santos\textsuperscript{1}}

% KAO: Remove extra newline
%\address{}{}{}
 \address{
	\textsuperscript{1 }Department of Computer Science, Universidade Federal de Minas Gerais, Belo Horizonte,  MG, Brazil - CEP 31270-901 \\
	\textsuperscript{2 }Universidade Cat\'olica de Bras\'ilia and Tribunal de Contas da Uni\~ao, Bras\'ilia, DF, Brazil - CEP 70042-900 \\
	\{edemirm, msalvim, jefersson\}@dcc.ufmg.br, matheusb@eng.grad.ufmg.br, remisb@tcu.gov.br
}

%\commission{VI, }{VI} %This field is optional.
%\workinggroup{VI/4} %This field is optional.
\icwg{}   %This field is optional.

\abstract{

In this work we present BrazilDAM, a novel public dataset based on Sentinel-2 and Landsat-8 satellite images covering all tailings dams cataloged by the Brazilian National Mining Agency (ANM). The dataset was built using  georeferenced images from 769 dams, recorded between 2016 and 2019. The time series were processed in order to produce cloud free images. The dams contain mining waste from different ore categories and have highly varying shapes, areas and volumes, making BrazilDAM particularly interesting and challenging to be used in machine learning benchmarks.  The original catalog contains, besides the dam coordinates, information about: the main ore, constructive method, risk category, and associated potential damage. To evaluate BrazilDAM's predictive potential we performed classification essays using state-of-the-art deep Convolutional Neural Network (CNNs). In the experiments, we achieved an average classification accuracy of 94.11\% in tailing dam binary classification task. In addition, others four setups of experiments were made using the complementary information from the original catalog, exhaustively exploiting the capacity of the proposed dataset.

%

%

%With the proposed novel dataset, we achieved an overall classification accuracy of 98.57\%. The resulting classification system opens a gate towards a number of Earth observation applications.

%We demonstrate how this classification system can be used for detecting land use and how it can assist in improving geographical maps. 
%\todo{Mencionar experimentos realizados.}

%The georeferenced dataset Brazil DAM will be made publicly available at \url{https://patreo.dcc.ufmg.br/datasets/brazildam} upon the acceptance of this paper.

%The geo-referenced dataset BrazilDAM will be made publicly available  upon the acceptance of this paper. 

%In this paper, we address the challenge of land use and land cover classification using Sentinel-2 satellite images. 

%The Sentinel-2 and Landsat-8 satellite images are openly and freely accessible provided in the Google Earth Engine Data Catalog.

%The resulting classification system opens a gate towards a number of Earth observation applications.
}

\keywords{Tailings Dam Detection, Remote Sensing, Deep Learning}

\maketitle

\section{Introduction}

On 25 January 2019 a tailings dam at the C\'orrego do Feij\~ao iron ore mine in Brumadinho, Brazil, suffered a catastrophic slope failure, followed by a mudflow that killed at least 248 people. This tragic event, three years and two months after the rupture of another large tailings dam in Mariana, which killed 19 people and destroyed the village of Bento Rodrigues, has resurrected the ghost of disasters that precarious structures can cause.

A tailings dam is typically an embankment used to store by products of mining operations. Tailings can be liquid, solid, or a mixture of fine particles suspended in liquid, usually toxic and potentially radioactive. Solid tailings are often used as part of the structure itself. These impoundments are designed for permanent containment and are between the largest man-made structures on Earth \cite{morgenstern2001geotechnics}.  

The number of tailings dam failures has doubled in the past 20 years. Advances in mining technology have made it possible to exploit lower grade deposits despite decreasing commodity prices, which means disposing of more rejects and putting more pressure on tailings facilities \cite{armstrong2019have}. Failure of tailings dams can be catastrophic, rapidly releasing large amounts of water and solid material, potentially causing large loss of life and huge damages to the environment and property. The risks, the challenges for long-term containment and the relatively poor safety-record revealed by the numbers of failures in tailings dams have led to an increasing awareness of the need for enhanced safety provisions~\cite{icold2001tailings}. 

There is no complete inventory of active tailings impoundments around the world.  The lack of any comprehensive tailings dam database has prevented meaningful analysis of the technical failures that could help prevent future incidents. The records are very incomplete on crucial data elements: design height of dam, design footprint, construction type (upstream, downstream, center line), age, design life, construction status, ownership status, capacity, release volume, runout, etc~\cite{yilmaz2017paste}.
Risk management and early-warning of dangerous trends, like accelerating displacements of slopes, are essential to support decision making, but require frequent high quality data. 

Remote-sensing has been increasingly used to build monitoring applications. The identification of precursors to catastrophic slope failures on tailing dams from space was proved possible using Interferometric Synthetic Aperture Radar (InSAR) \cite{Carla2019}. Satellite images can also be used in combination with other sources of information in order to assess risk or investigate specific circumstances on tailings dams collapses \cite{williams1999review}.
Remote-sensing image analysis in combination with machine learning methods have seen a massive rise in popularity for  over the past few years. Machine learning has been applied to tasks including image fusion, image registration, scene classification, object detection, land use and land cover (LULC) classification, segmentation, and object-based image analysis (OBIA) \cite{MA2019166}.
The use of these combined technologies can be very useful for the investigation of issues concerning the tailings dams. Scene classification and image segmentation could be used to acquire comprehensive tailings dam databases, including footprints, heights, volumes, changes throughout time and contained waste classification. These databases could be used for risk assessment and monitoring by regulatory agencies and local communities.

Machine learning methods require rich training datasets in order to fit, evaluate and test predicting models. The acquisition and preparation of these datasets can be arduous or even impracticable if there are no trustful sources of ground truth information.

In order to facilitate and stimulate the use of machine learning methods on the research on tailings dams issues, we have built a dataset, named BrazilDAM, based on Sentinel-2 and Landsat-8 satellite images covering all tailings dams cataloged in Brazil.

%In this paper, we present our dataset\footnote{\url{XXX}}, \textbf{that will be made publicly available upon the acceptance of this paper}, and show preliminary results on the use of deep learning methods for the discovery of tailings dams on large area satellite images.

In this paper, we present our dataset\footnote{\url{http://www.patreo.dcc.ufmg.br/brazildam-dataset/}}, and show preliminary results on the use of deep learning methods for the discovery of tailings dams on large area satellite images.

\subsection{Motivations}

Brazil is one of the main mineral exporters in the world. Mining accounts for almost 7\% of Brazil's GDP and generates hundreds of thousands of jobs, making it one of the most important aspects of national development and economic stability.
Despite its importance, mining activity has caused severe environmental impact. Among the different environmental issues, mining activities has caused landscape degradation, erosion, soil contamination, groundwater and surface water pollution~\cite{IBRAM2017}. 
According to a report by the Brazilian National Water Agency - ANA \footnote{https://www.ana.gov.br/noticias/45-barragens-preocupam-orgaos-fiscalizadores-aponta-relatorio-de-seguranca-de-barragens-elaborado-pela-ana/rsb-2017.pdf/view ; Accessed 29-November-2019}, there are 780 mining tailings dams in Brazil, which are part of the 24,092 dams cataloged by the agency. 

According to the report released by ANA, referring to 2017, only 42\% of the known dams are officially licensed. The lack of comprehensive information forbids the proper risk classification of 76\% of all dams. The inspection of the dams is responsibility of 39 different regional and four Brazilian federal government agencies. However, most of regional agencies have no official assigned team to work on dam safety control. This is the case of state of Minas Gerais, where the most recent  tragedies occurred. As a consequence, in 2017 only 3\% of the all catalogued dams were visited by the supervisory bodies. Compromised structures were detected in 45 of the visited dams.
ANA centralizes the data reported by the inspection bodies and is supposed to maintain and share the government official dams database. Nevertheless, the lack of vital information is evident. For instance, 18,446 dams have no height information, 9,584 have no capacity information, and 18,663 dams have not been classified for their potential damage. Without this information it is not possible to identify risky dams that should be closely monitored. There is also evidence of a probably large number of uncatalogued dams. In fact, Minas Gerais State Secretariat of Environment and Sustainable Development - SEMAD, responsible for auditing dams in that state, reported in 2017 the existence of 57 dams, a number much lower than the 698 structures registered in the State of Minas Gerais Environmental Foundation Dam Database. 

The depicted scenario points to the need for better quality information and better monitoring instruments concerning the tailings dams in Brazil. The recent technologies on remote sensing and machine learning can be used to create helpful solutions to be used by the public administration and its supervisory bodies but also by local communities and non governmental organizations.   

\subsection{Challenges and Contributions}
%\subsection{Contributions}

In this paper, we claim the following contributions:
(1) we introduce a novel public dataset based on Sentinel-2 and Landsat-8 satellite images covering all tailings dams cataloged in Brazil; and (2) we provide a benchmark addressing the challenge of land-use and land-cover classification for the proposed BrazilDAM dataset using several modern CNN architectures.

%\begin{enumerate}

 %   \item We introduce a novel public dataset based on Sentinel-2 and Landsat-8 satellite images covering all tailings dams cataloged in Brazil.
    
  %  \item We provide a benchmark addressing the challenge of land-use and land-cover classification for the proposed BrazilDAM dataset using several modern CNN architectures.
    %\msa{Are you naming your datase ``BrazilDAM''? If so you should make it clear that's its name and use it consistently throughout the paper (even in the abstract).}
%\end{enumerate}

\section{Dataset Acquisition}
    %On November 5th, 2015 and January 25th, 2019 ore rejects dams collapsed in Brazil. The rupture causes several problems, both social and economics \cite{freitas2019samarco}. After that, the responsible division of government published a table\footnote{http://www.anm.gov.br/assuntos/barragens/pasta-cadastro-nacional-de-barragens-de-mineracao/classificacao-oficial-anm} with some data about registered dams. There is some regulatory information like name, company name, national register of legal entities, latitude, longitude, locate, country, city and if it is subscribed in PNSB (\textit{Política Nacional de Segurança de Barragens})  that is a public policy to ensure the integrity and try to prevent future problems.  There is also some technical information that was made by experts like main ore, height, volume, constructive method, risk category, potential damage associated. All of this data coupled, with the availability of satellite imagery, provides an enabling environment for specialized dataset creation.
    
    %sugestao de nova redacao
    
    The Brazilian National Mining Agency makes available on its website a georeferenced database describing all officially registered tailings dams in the country \cite{anm-info}.
    %\footnote{http://www.anm.gov.br/assuntos/b%arragens/pasta-cadastro-nacional-de-barrag%ens-de-mineracao/classificacao-oficial-anm}. 
    The database contains some identification information about the dams and their owners and a spatial coordinates of a point indicating each dam's location. There is also some technical information like the main ore, height, volume, constructive method, risk category, potential damage associated with the dam. 

    %It is important to emphasize that NASA and the European Space Agency (ESA) steps up efforts to improve Earth observation, with the Copernicus program and the Landsat Mission. Along with Google Earth Engine \cite{gorelick2017google} that encapsulates the amount of data and the processing, allow us to acquire multi-spectral images with a temporal component since June 2015 (when the Sentinel-2 satellite was successfully launched). 
    
The main providers of open access satellite imagery are NASA and the European Space Agency (ESA), with the Copernicus-Sentinel program and the Landsat Mission respectively. Both of the satellites used, Sentinel-2 and Landsat 8, are sun-synchronous. They capture images with Multispectral Imager (MSI), however, exist differences between the resolution and the wavelength of each band.

% NASA's Landsat program provides the longest continuous space-based record of Earth’s land in existence. Since the early 1970s, Landsat has continuously and consistently archived images of Earth. Sentinel-2 is an Earth observation mission from the Copernicus Programme that systematically acquires optical imagery at high spatial resolution (10 m to 60 m) over land and coastal waters.  

To deal with Landsat and Sentinel-2 image collections we chose to use the Google Earth Engine platform. The Google Earth Engine (GEE) \cite{gorelick2017google} is a cloud computing platform designed to store and process huge datasets. The easily accessible and user-friendly front-end provides a convenient environment for interactive data and algorithm development. Google archived all the Landsat and Sentinel image collection and linked them to the cloud computing engine for open source use. 
Besides to provide the computational infrastructure and the image collections, the GEE API allows the images processing to deal with some commons problems, as cloud cover.

 According to \cite{esa-radiometric} the 13 spectral bands of Sentinel-2 range from the Visible (VNIR) and Near Infra-Red (NIR) to the Short Wave Infra-Red (SWIR). Also has three bands that measure atmospheric effects (aerosols, cirrus, and water vapor). Sentinel's resolutions for RGB are 10 meters per pixel. This satellite can cover the entire world in five days, and generate a temporal ax of multi-spectral images.
 
%  Where this resolution is radiometric, and measure the capacity of the instrument to distinguish differences in light intensity or reflectance. The greater the radiometric resolution, the more accurate the sensed image will be.  
 
 Still,  according to \cite{landsat-info},  the 11 spectral bands of Landsat 8 are collected in a different resolution (30 meters per pixel in RGB) and wavelength range, although there is an intersection between then. Landsat 8 can cover the entire globe every sixteen days, doing 223 orbit cycle.
 
All this information available freely allows us to create a new dataset that covers all dam labeled. Was used all bands of both satellites and an annual time-series from January 1st, 2016 until September 5th, 2019 (the date that our algorithm starts running).

\subsection{Satellite Image Acquisition}

To deal with this amount of data of both satellites, we choose to use the Google Earth Engine. The engine works with a data structures based on requests, the Image Collection is a stack of images based in Earth Engine collection ID. This structure allows operations like filtering, mapping, reducing, compositing and iterating. Regarding the Image structure, are a raster composed of one or more bands that withhold their name, data type, scale, mask, and projection.

% This platform is a cloud-based geospatial processing for scientific analysis and visualization. Besides to provide the computational infrastructure, the API allows pre-processing the images to manager some commons problems, as cloudy places like \textit{Amazônia} where exist several ore dams. 

% The engine works with a data structures based on requests, the Image Collection is a stack of images based in Earth Engine collection ID. This structure allows operations like filtering, mapping, reducing, compositing and iterating. Regarding the Image structure, are a raster composed of one or more bands that withhold their name, data type, scale, mask, and projection.

\begin{figure}[h!]
    \includegraphics[width=1.\linewidth]{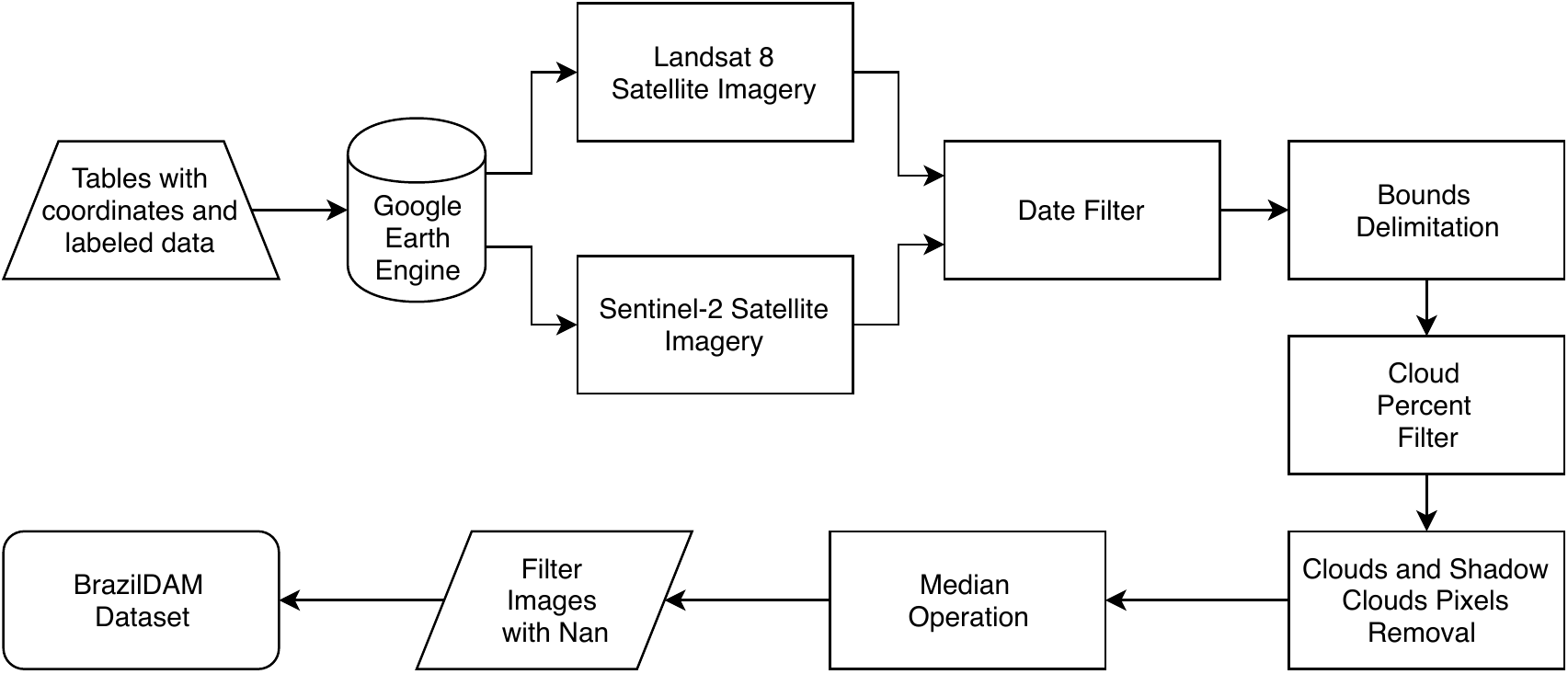}
    \centering
    \caption{Workflow that illustrates the image acquisition.}
    \centering
    \label{fig:workflow}
\end{figure}

For image acquisition, we use the \textit{COPERNICUS/S2} Image Collection for Sentinel and \textit{LANDSAT/LC08/C01/T1\_TOA} for Landsat. As can be seen a workflow, in Figure \ref{fig:workflow}, after choosing the satellite some filters were made to improve the overall quality.

% To improve the quality of the images, we run a function propose in the API. It consists in removing the pixels of all images in the Image Collection that are surely clouds and shadows of clouds. After that, got the median of those, creating a unique image representing the whole year.

The years were separated into four different folders, using the date filter images of the years 2016, 2017, 2018 and 2019 were obtained. 
This was done to create the time axis in the dataset.

Then, a boundary filter delimits the context of image acquisition. A square geometry was made, leaving the point coordinate exactly in the center of the image. The distance between the center and the closest borders are 1.9 kilometers, this resulted in images with 384x384 pixels.

Using the metadata of the images, it was also possible to filter the percentage of clouds within the images. This makes possible to present the features more easily. 

To improve the quality, a function was mapped in each of the images of the collection, which removed all the pixels that were marked as a cloud by the atmospheric context bands (QA60 for the Sentinel and BQA in Landsat). Finally, to return a raster Image was made a median operation between the images of the Image Collection.

This operation, depicted in Equation \ref{equ:1}, is based on the assumption that clouds are not in the same place permanently. Therefore, most parts of pixels in the same region will not have clouds. In making the median we look for the most common pixel value, which will be closest to the actual pixel value in that region. This operation is done individually in each of the bands.

\begin{equation}\label{equ:1}
    \rho(x_j) \approx median(\rho(x_{1, 2, 3, ..., k}))
\end{equation}

\begin{tabbing} 
where \hspace{0.6cm} \= $\rho(x_j)$ = the observed band of interest\\
\> $x$ = image with year delimitation\\
\> $j$ = the j\textit{th} pixel\\
\> $k$ = is the total number of pixels 
\end{tabbing}

With all the images of that year, the demarcated context area, no cloud pixels and the median operation performed are generate the final image. However, we face some problems. By filtering the amount of cloud by lest then 30\%, some of the collections returned no images. Some regions in Brazil are very humid and have clouds for much of the year. To deal with this limitation, we made an algorithm that increases this percentage by 10\% until they're at least one image.

Another problem was some medians returning NaN (not a number). This was due to the removal of the flagged pixels with clouds, all pixels on that time axis were removed. For these cases, the median was calculated without the cloud removal function. Even so, some requests did not return any images. In this case, some folders were left with fewer files.

\begin{figure}[h]
\centering
\subfigure[Not dam acquisition.]{
    \includegraphics[width=.475\columnwidth]{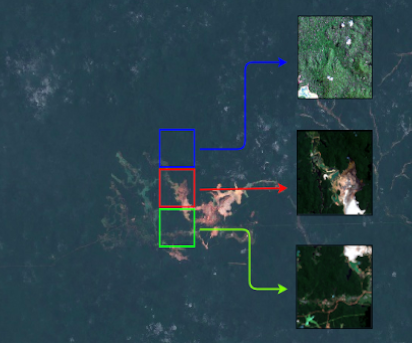}
    \label{fig:not_dam_explanation}
}
\subfigure[Intersection issue.]{
    \includegraphics[width=.475\columnwidth]{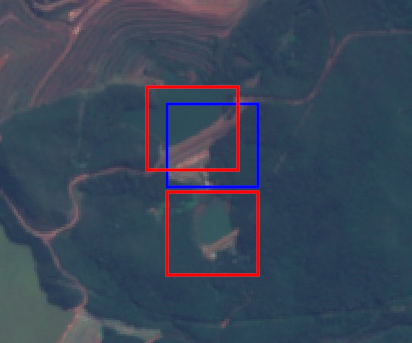}
    \label{fig:intersection}
}
\caption[]{Explanation of not dam images acquisition and intersection issues.}
\label{fig:dataset_acquisition}
\end{figure}

For the not-dam locations, where was used the coordinate above and under each clip. This can be seen in Figure \ref{fig:not_dam_explanation}, where the red square is the clip of a dam and in blue and green are the clips of not dams. To ensure that none of the not dam label contains a dam in itself, we delete the images where exists an intersection with the base location of the dam as can be noticed in Figure \ref{fig:intersection}.

%  \begin{figure}[ht]
%     \includegraphics[width=\linewidth]{images/get_dam_and_not_dam.png}
%     \centering
%     \caption{Explanation to how to get the imagens of not dams. In red is represented the imagem of a dam labeled in our dataset, in blue is show a image of not dam in north and in green in south. }
%     \centering
%     \label{fig:not_dam_explanation}
% \end{figure}

% \begin{figure}[h]
%     \includegraphics[width=\linewidth]{images/intersection_.png}
%     \centering
%     \caption{In red is showed the images of dams in blue we can see a image of not dam the intersect a dam image, in this case the image in blue not was added to our dataset.}
%     \centering
%     \label{fig:intersection}
% \end{figure}

% Figure \ref{fig:not_dam_explanation} explains how to get the images of not dams. In red is represented the image of a dam labeled in our dataset, in blue is shows an image of not dam in the north and green in the south.  Also in Figure \ref{fig:intersection}, the red is showed the images of dams in blue we can see an image of not dam intersect a dam image, in this case, the image in blue not was added to our dataset.

\begin{figure}[h]
\centering
\subfigure[High noise image. ]{
    \includegraphics[width=.475\columnwidth]{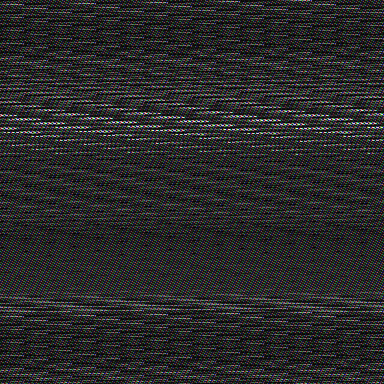}
    \label{fig:subfig1}
}
\subfigure[Mosaic image.]{
    \includegraphics[width=.475\columnwidth]{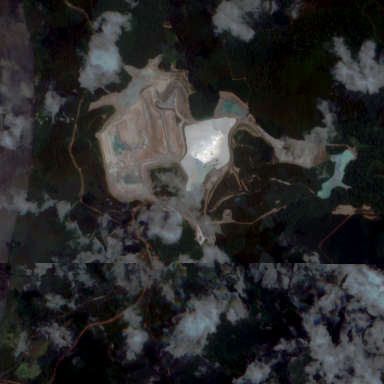}
    \label{fig:subfig2}
}
\caption[]{Examples of bad images in acquisition of dams.}
\label{fig:badimages}
\end{figure}

It is important to attend that some images get noised, or neither were added to the dataset. The reasons of those were the impression of data coordinates, few quality images in a year or artifacts created by the median operation. As are shown in Figure \ref{fig:badimages}, where the \ref{fig:subfig1} there is an example of failure in the satellite acquisition and in \ref{fig:subfig2} we have some artifacts created by the median when a mosaic was necessary.
% The final version of the dataset obtained the number of images as shown in Table \ref{tab:info-dataset}.

% \begin{table}[h!]
% \centering
% \begin{tabular}{|l|c|c|c|c|}
% \toprule
% Satellite & \multicolumn{2}{c|}{Landsat 8} & \multicolumn{2}{c|}{Sentinel 2} \\ \midrule
% Year      & Dam          & Not Dam         & Dam          & Not Dam          \\ \hline
% 2016      & 761          & 1143            & 769          & 1156             \\ 
% 2017      & 769          & 1155            & 769          & 1156             \\ 
% 2018      & 769          & 1156            & 769          & 1156             \\ 
% 2019      & 756          & 1130            & 769          & 1156             \\ \bottomrule
% \end{tabular}
% \caption{Specifying the amount of images in the dataset.}
% \label{tab:info-dataset}
% \end{table}

% % Please add the following required packages to your document preamble:
% % \usepackage{booktabs}
% \begin{table}[]
% \centering
% \begin{tabular}{@{}ccccc@{}}
% \toprule
% Satellite & \multicolumn{2}{c}{Landsat 8} & \multicolumn{2}{c}{Sentinel 2} \\ \midrule
% Year & Dam & Not Dam & Dam & Not Dam \\ \midrule
% 2016 & 761 & 1143 & 769 & 1156 \\
% 2017 & 769 & 1155 & 769 & 1156 \\
% 2018 & 769 & 1156 & 769 & 1156 \\
% 2019 & 756 & 1130 & 769 & 1156 \\ \bottomrule
% \end{tabular}
% \caption{Specifying the amount of images in the dataset.}
% \label{tab:info-dataset}
% \end{table}
\section{Dataset Benchmarking}

%In order to establish a benchmark for the new dataset, we proposed a set of experiments using state-of-the-art deep learning networks. for the classification task.

%In order to establish a benchmark for the proposed dataset, a set of experiments was done using state-of-the-art deep learning networks in the following tasks: Dam classification, Ore classification, Constructive method, Risk category and Associated Potential Damage.

To establish a benchmark for the proposed dataset, experiments was done using state-of-the-art deep learning networks in the following tasks: Dam classification, Ore classification, Constructive method, Risk category and Associated Potential Damage.

% \begin{enumerate}
%     \item Dam classification: 
    
%     Given aerial images of regions divided into two classes, dams and non-dams, how to assign the correct label in a supervised manner using an image classification network.
    
%     \item Ore classification: 
    
%     Given aerial images containing regions mining dams. How to discriminate and assign different types of mining supervised among images using image classification.
    
%     \item Constructive method:

%     \item Risk category:

%     \item Associated Potential Damage:
    
% \end{enumerate}{}

%Para cada uma das tarefas realizadas, foram utilizadas apenas a imagens extraidas no ano de 2019. Sem perda de generalidade os resultados e discussoes podem ser extendidos para o restante dos dados. Afim de explorar máximo os dados, em cada uma das tarefas de classificação foram utilizados seis diferentes arquiteturas de redes neurais: alexnet, densenet, inception, resnet, squeezenet e vgg. Para o treinamento dos modelos foi utilizada a implementação da biblioteca pytorch . Além disso foi utilizado o algoritmo Adam como otimizador em cada uma das redes, variando o parametro de taxa de aprendizado entre $\lambda = {10^-2,10^-3,10^-4}$.
For each of the tasks performed, only images extracted from sentinel satellite in the year 2019 were used. To statistically validate the classification results, the cross-validation procedure was used, where the data used for training and testing are rotated to simulate the results in an independent data set. Without loss of generality, the results and discussions can be extended to the rest of the data. In order to fully exploit the data, six different neural network architectures were used for each classification task: AlexNet \cite{krizhevsky2014one}, DenseNet \cite{huang2017densely}, Inception \cite{szegedy2016rethinking}, ResNet \cite{he2016deep}, SqueezeNet \cite{iandola2016squeezenet} and VGG \cite{simonyan2014very}. For the training of the models the implementation of the PyTorch library \cite{paszke2017automatic} was used. In addition, the Adam algorithm was used as optimizer in each of the networks, varying the learning rate parameter between $ \lambda = \{10^{-2},10^{-3},10^{-4}\}$. Reported results were acquired by averaging the balanced accuracy across the 5 cross-validation folds. Besides, 95\% confidence intervals of the mean were also reported. In ours report tables, the first column contains the methods and learning rate that achieved the best results. The second and third columns show a mean balanced accuracy and 95\% confidence interval obtained from 5-fold cross-validation. Each of the following sections details the experimental protocol used to evaluate the proposed tasks, as well as the results,  obtained followed by a discussion of it.

\subsection{Dam Classification}

%\ede{Descrição inicial do problema}

According to ANA, a large part of the ore dams have not yet been registered or cataloged in the national safety policy. This is expensive work and requires a lot of manpower to cover the entire Brazilian territory. While there is a duty for companies to report dam construction, they often do not, causing soil contamination or disruption due to poor planning.

%According to ANA, the Brazilian National Water Agency, a large part of the ore dams have not yet been registered or cataloged in the national safety policy. This is expensive work and requires a lot of manpower to cover the entire Brazilian territory. While there is a duty for companies to report dam construction, they often do not, causing soil contamination or disruption due to poor planning.

% %https://www.ana.gov.br/noticias/45-barragens-preocupam-orgaos-fiscalizadores-aponta-relatorio-de-seguranca-de-barragens-elaborado-pela-ana

%ler depois
%Using the detection algorithm shown in this article, it is possible to make these records remotely and much faster with 92\% accuracy in the worst case. 

%Na tarefa de classificação de barragens, foram utilizadas no total 1925 imagens (onde 769 são rotuladas como barragens e 1156 como não barragens). De forma a validar estatisticamente os resultados da classificação foi utilizado o procedimento de validação cruzada, onde os dados utilizados para treino e teste são rotacionados de forma a simular os resultados em um conjunto de dados independentes.

In the dam classification task, a total of 1925 images were used (where 769 are labeled as dams and 1156 as non-dams). The results obtained can be seen in the Table \ref{tab:dam_result}

% \begin{table}[h]
% \caption{Dam Classification}
% \tabcolsep=0.155cm
% \begin{tabular}{ccccc}
% \hline
% \textbf{Config}        & \textbf{Mean}   & \textbf{CI}     & \textbf{CI\_MIN} & \textbf{CI\_MAX} \\ \hline
% vgg\_0.0001            & 0.9323          & 0.0157          & 0.9165           & 0.9480           \\ \hline
% inception\_0.0001      & 0.9361          & 0.0159          & 0.9202           & 0.9520           \\ \hline
% \textbf{dense\_0.0001} & \textbf{0.9411} & \textbf{0.0181} & \textbf{0.9230}  & \textbf{0.9592}  \\ \hline
% squeeze\_0.0001        & 0.8944          & 0.0287          & 0.8657           & 0.9231           \\ \hline
% alexnet\_0.0001        & 0.9177          & 0.0231          & 0.8946           & 0.9409           \\ \hline
% resnet\_0.001          & 0,9078          & 0.0135          & 0.8943           & 0.9213           \\ \hline
% \end{tabular}
% \label{tab:dam_result}
% \end{table}

% Please add the following required packages to your document preamble:
% \usepackage{booktabs}
\begin{table}[h!]
\centering
%\caption{Dam Classification}

\begin{tabular}{@{}ccc@{}}
\toprule
\textbf{Config}         & \textbf{A.A. (\%)} & \textbf{CI 95\%}       \\ \midrule
VGG 1E-04               & 93.23            & 91.65 - 94.80          \\
Inception 1E-04         & 93.61            & 92.02 - 95.20          \\
\textbf{DenseNet 1E-04} & \textbf{94.11}   & \textbf{92.30 - 95.92} \\
SqueezeNet 1E-04        & 89.44            & 86.57 - 92.31          \\
AlexNet 1E-04           & 91.77            & 89.46 - 94.09          \\
ResNet 1E-03            & 90.78            & 89.43 - 92.13          \\ \bottomrule
\end{tabular}
\caption{This table shows the balanced accuracy of methods on the Dam Classification Task.}
%\caption{This table shows the balanced accuracy of methods on the Dam Classification Task. The first column contains the methods and learning rate that achieved the best results. The second and third columns show a mean balanced accuracy and 95\% confidence interval obtained from 5-fold cross-validation}
\label{tab:dam_result}
\end{table}

\begin{figure}[h!]
\subfigure[False Negative]{
    \includegraphics[width=.475\columnwidth]{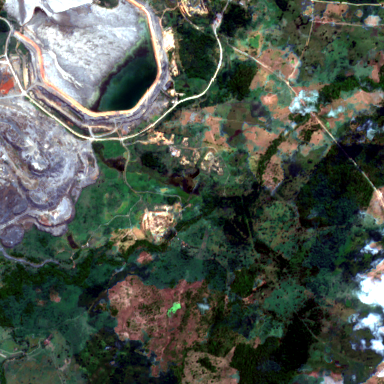}
    %\label{fig:subfig1}
}
\subfigure[False Positive]{
    \includegraphics[width=.475\columnwidth]{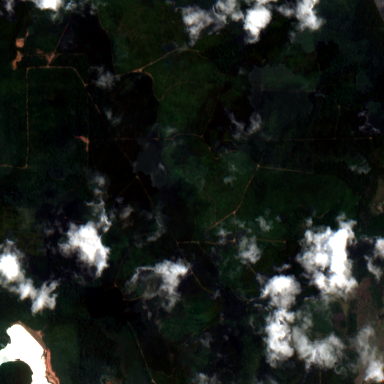}
    %\label{fig:subfig4}
}
\caption[]{Examples of dam and non-dam samples incorrectly labeled. Images (a) is false negative (dam image predicted as non-dam) and image (b) is a false positive (non-dam image that have been classified as dam)}
\label{fig:dam_misses}
\end{figure}

As can be observed in the results, the model with the largest lower level of the confidence interval was highlighted. Despite the result obtained by DenseNet, all the architectures used had results considered satisfactory by the authors. %In fact, there is a large intersection between the reported confidence intervals.
In order to better understand the misclassifications of the generated model, we highlight two examples of incorrectly classified samples. Examples of false positive and negative can be seen in Figure \ref{fig:dam_misses}.

As can be seen from figure \ref{fig:dam_misses}, image (a) show an example of false negative (image that is a dam but was classified as non-dam). In figure (a), although it is possible to observe a dam in the upper left corner of the image, its center is composed only of vegetation. This is a case of geographic coordinate error, where data provided by ANA do not correctly present the center of the structure. In the example of image (b), the image have poor visual quality, with the presence of artifacts or clouds.

%\ede{Esse resultado é o suficiente para possibilitar monitorar em larga escala, pela presença de barragens ilegais em diferentes lugares do brasil.}

\subsection{Ore classification}

%Barragens de rejeito são estruturas de terra construídas para armazenar resíduos de mineração. Esses residuos são definidos como a fração estéril produzida pela extração de minérios, em um processo mecânico e/ou químico que divide o mineral bruto em concentrado e rejeito. Apesar do rejeito ser um material que não possui grande valor econômico, o seu armazenamento é crucial para evitar impactos socioambientais.

Tailings dams are earth structures built to store mining waste. These residues are defined as the sterile fraction produced by ore extraction in a mechanical and/or chemical process that divides the crude mineral into concentrate and tailings. Although the waste is a material that does not have great economic value, its storage is crucial to avoid social and environmental impacts. In this task, the dam images were classified according to their main mining tailings. This monitoring is important to ensure that no dams are receiving deposits of other types of tailings illegally.

%In this experiment, a total of 769 images of dams were used. Dentre as 769 barragens existem um total de 59 diferentes tipos de minerios. Assim para evitar o desbalanceamento entre as classes, foram selecionadas apenas as classes com pelo menos 10 amostras, resultando em 15 classes. O restante das imagens foram atribuidas para a classe "outros", totalizando 16 classes.

%Dado que nessa tarefa é feito uma classificação multiclasse, para uma melhor interpretação dos resultados podemos utilizar a matrix de confusão vista na figura 7.

In this experiment, a total of 769 images of dams were used. Among the 769 dams, there are a total of 59 different types of ores. Thus to avoid unbalance between classes, only classes with at least 10 samples were selected, resulting in 15 classes. The rest of the images were assigned to the "others" class, totaling 16 classes. The results obtained can be seen in Table \ref{tab:ore_result}. Since in this task a multiclass classification is made, for a better interpretation of the results we can use the confusion matrix seen in Figure \ref{tab:ore_result}.

% \begin{table}[]
% \caption{Ore Classification}
% \tabcolsep=0.155cm
% \begin{tabular}{ccccc}
% \hline
% \textbf{Config}        & \textbf{Mean}   & \textbf{CI}     & \textbf{CI\_MIN} & \textbf{CI\_MAX} \\ \hline
% vgg\_0.0001            & 0.6022          & 0.0608          & 0.5414           & 0.6630           \\ \hline
% inception\_0.0001      & 0.6913          & 0.0386          & 0.6527           & 0.7299           \\ \hline
% \textbf{dense\_0.0001} & \textbf{0.7242} & \textbf{0.0387} & \textbf{0.6855}  & \textbf{0.7630}  \\ \hline
% squeeze\_0.0001        & 0.6371          & 0.0407          & 0.5964           & 0.6779           \\ \hline
% alexnet\_0.0001        & 0.5796          & 0.0603          & 0.5192           & 0.6399           \\ \hline
% resnet\_0.0001         & 0.7162          & 0.0324          & 0.6838           & 0.7487           \\ \hline
% \end{tabular}
% \end{table}

\begin{table}[h!]
\centering
\begin{tabular}{@{}lcc@{}}
\toprule
\multicolumn{1}{c}{\textbf{Config}} & \textbf{A.A. (\%)} & \textbf{CI 95\%}       \\ \midrule
VGG 1E-04                           & 60.22            & 54.14 - 66.30           \\
Inception 1E-04                     & 69.13            & 65.27 - 72.99          \\
\textbf{DenseNet 1E-04}             & \textbf{72.42}   & \textbf{68.55 - 76.30} \\
SqueezeNet 1E-04                    & 63.71            & 59.64 - 67.79          \\
AlexNet 1E-04                       & 57.96            & 51.92 - 63.99          \\
ResNet 1E-04                        & 71.62            & 68.38 - 74.87          \\ \bottomrule
\end{tabular}

\caption{This table shows the balanced accuracy of methods on the Ore Classification Task.}
%\caption{This table shows the balanced accuracy of methods on the Ore Classification Task. The first column contains the methods and learning rate that achieved the best results. The second and third columns show a mean balanced accuracy and 95\% confidence interval obtained from 5-fold cross-validation}
\label{tab:ore_result}
\end{table}

\begin{figure}[h]
    \includegraphics[width=\columnwidth]{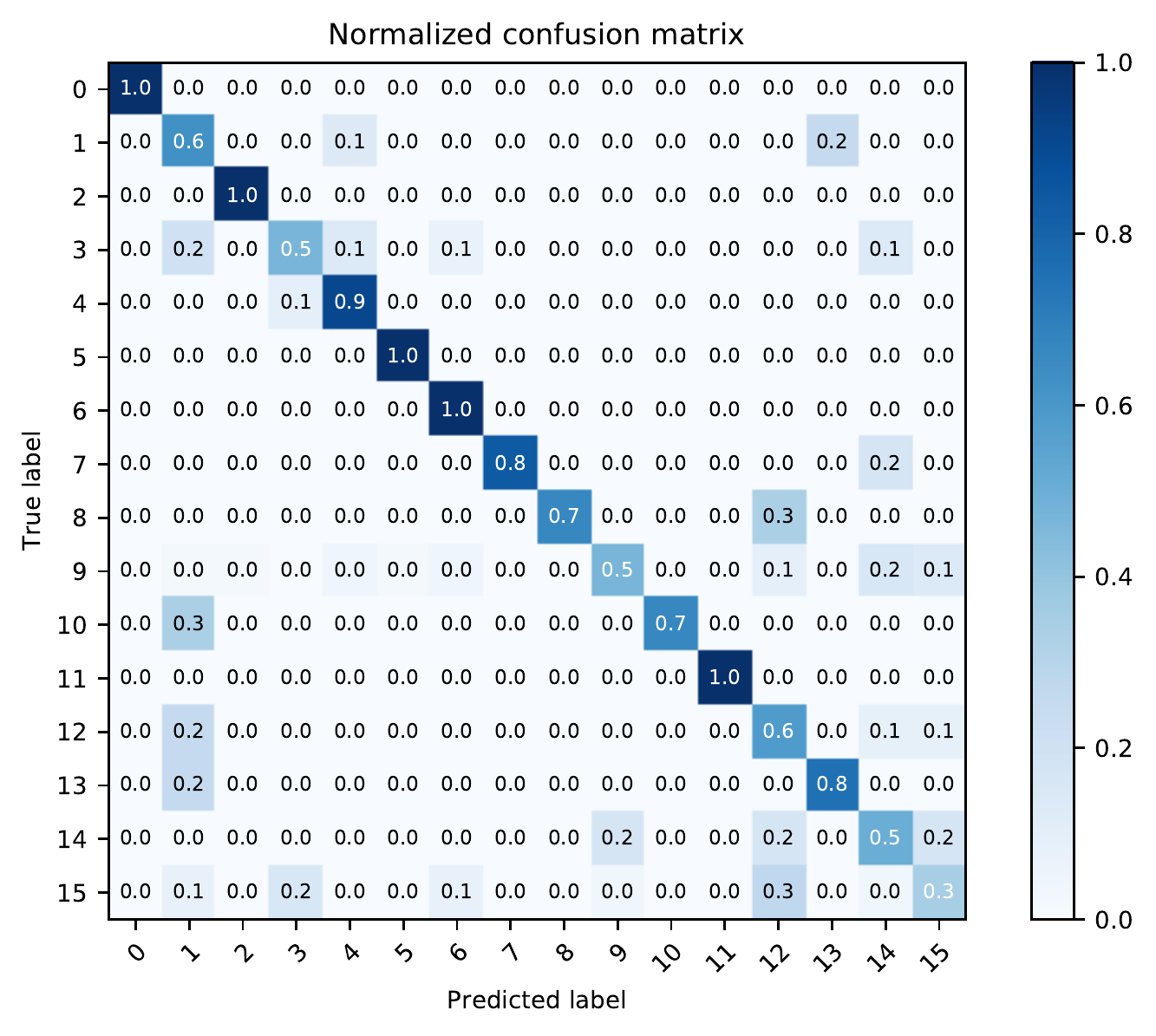}
    %\label{fig:subfig1}
\caption{Confusion matrix of Ore Classification task. %\caption{Confusion matrix of Ore Classification task. Most misclassifications occurred for results classified the "Others" category. %Adding "others" category reduced the number of misclassifications into , but significantly increased the number of images in the Other category categorized as Diagnostic. 
}
\label{fig:ore_cm}
\end{figure}

%Observando a Tabela 4, é possivel observar novamente que o melhor resultado foi obtido pela densenet atingindo cerca de 72% da acuracia balanceada. Entretanto, como podemos ver na matrix de confusão da figura 7, com exceção da classe 15, que represença a categoria "others", o modelo conseguiu discriminar com sucesso os diferentes tipos de minerios entre as imagens de barragem, considerando que em um cenario multiclasse, o resultado aleatorio é de 6%.

Looking at Table \ref{tab:ore_result}, it is possible to observe again that the best result was obtained by DenseNet reaching about 72\% of the balanced accuracy. However, as we can see in the confusion matrix of Figure \ref{fig:ore_cm}, except for class 15, which represents the category "others", the model was able to successfully discriminate the different types of minerals between the dam images, considering that in a multiclass scenario, the random result is 6\%.

%\ede{Esse resultado é o suficiente para adquirir de tempos em tempos imagens e monitorar se o minério que esta sendo despejado ainda continua o mesmo OU verificar se novas barragens de fato estão despejando o minério 'combinado'}

\subsection{Construction Methods}

%A construção de barragens de rejeito são influenciadas diretamente pelo tipo de rejeito a ser depositado. Barragens por aterro hidráulico pode ser feita a partir três métodos principais: alteamento à montante, alteamento à jusante e método da linha de centro. Apesar de todas as formas partirem da construção de um dique de partida, a diferença dos métodos se da principalmente na direção em que o alteamento é feito. O monitoramento do método utilizado na sua construção é extremamente importante para evitar a criação de barragens construidas de forma ilegal, que não sejam adequadas para uma determinada area ou determinado deposito de minerio.

%The construction of tailings dams is directly influenced by the type of tailings to be deposited. Hydraulic embankment dams can be made from three main methods: upstream elevation, downstream elevation, and centerline method. Although all forms start from the construction of a starting dike, the difference in methods is mainly in the direction in which the elevation is made. The monitoring method used in its construction is extremely important to avoid the creation of dams built illegally, which are not suitable for a particular area or particular ore deposit.

The construction of tailings dams is directly influenced by the type of tailings to be deposited. Hydraulic embankment dams can be made from three main methods: upstream elevation, downstream elevation, and centerline method. Although all forms start from the construction of a starting dike, the difference in methods is mainly in the direction in which the elevation is made. The monitoring method used in its construction is extremely important to avoid the creation of dams built illegally, which are not suitable for a particular area or ore deposit.

%In this experiment, a total of 769 images of dams were used. Dentre as 769 barragens existem um total de 5 diferentes tipos de construções: Downstream method, Upstream method or unknown, Center Line method, Single step method, Undefined

In this task, a total of 769 images of dams were used. Among the 769 dams, there are a total of 5 different building types: Downstream method, Upstream method or unknown, Center Line method, Single-step method, Undefined. The results obtained can be seen in the Table \ref{tab:construction_result}

% \begin{table}[h!]
% \caption{Constructive Method}
% \label{tab:construction_result}
% \tabcolsep=0.155cm
% \begin{tabular}{ccccc}
% \hline
% \textbf{Config}        & \textbf{Mean}   & \textbf{CI}     & \textbf{CI\_MIN} & \textbf{CI\_MAX} \\ \hline
% vgg\_0.0001            & 0.5118          & 0.0215          & 0.4903           & 0.5333           \\ \hline
% inception\_0.0001      & 0.5330          & 0.0250          & 0.5081           & 0.5580           \\ \hline
% \textbf{dense\_0.0001} & \textbf{0.5547} & \textbf{0.0507} & \textbf{0.5041}  & \textbf{0.6054}  \\ \hline
% squeeze\_0.0001        & 0.4389          & 0.0586          & 0.3804           & 0.4975           \\ \hline
% alexnet\_0.0001        & 0.4916          & 0.0319          & 0.4597           & 0.5235           \\ \hline
% resnet\_0.0001         & 0.4525          & 0.0115          & 0.4410           & 0.4640           \\ \hline
% \end{tabular}
% \end{table}

% Please add the following required packages to your document preamble:
% \usepackage{booktabs}
\begin{table}[h!]
\centering
%\caption{Constructive Method}

\begin{tabular}{@{}ccc@{}}
\toprule
\textbf{Config}          & \textbf{A.A. (\%)} & \textbf{CI 95\%}       \\ \midrule
VGG 1E-04                & 51.18            & 49.03 - 53.33          \\
\textbf{Inception 1E-04} & \textbf{53.30}    & \textbf{50.81 - 55.80} \\
DenseNet 1E-04           & 55.47            & 50.41 - 60.54          \\
SqueezeNet 1E-04         & 43.89            & 38.04 - 49.75          \\
AlexNet 1E-04            & 49.16            & 45.97 - 52.35          \\
ResNet 1E-04             & 45.25            & 44.10 - 46.40          \\ \bottomrule
\end{tabular}
\caption{This table shows the balanced accuracy of methods on the Construction Method Task. %\caption{This table shows the balanced accuracy of methods on the Construction Method Task. The first column contains the methods and learning rate that achieved the best results. The second and third columns show a mean balanced accuracy and 95\% confidence interval obtained from 5-fold cross-validation. %\todo{Os valores estão no padrão brasileiro. Com vírgula.}
}
\label{tab:construction_result}
\end{table}

%
%\begin{figure}[h]
%    \includegraphics[width=\columnwidth]{images/dam_miss/Confusion_matri%x_construction.pdf}
%    %\label{fig:subfig1}
%\caption[]{}
%\label{fig:construction_cm}
%\end{figure}

\begin{figure}[h]
\subfigure[Upstream method]{
    \includegraphics[width=.475\columnwidth]{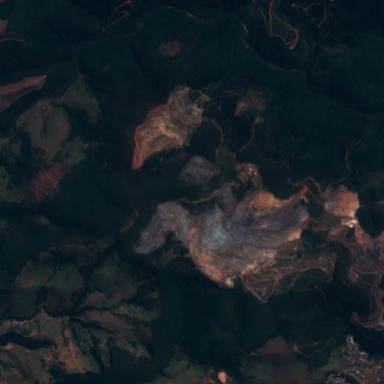}
    %\label{fig:subfig1}
}
\subfigure[Single Step method]{
    \includegraphics[width=.475\columnwidth]{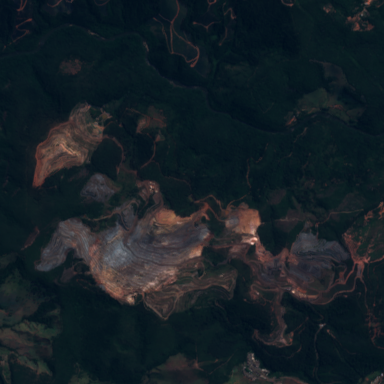}
    %\label{fig:subfig2}
}

%Exemplos de barragens de diferentes métodos de construção dentro da mesma regiao. Na imagem (a), apresenta uma barragem construida usando o método de Alteamento a Montante, enquanto em uma região proxima, temos a imagem (b) de uma barragem construida usando o método de etapa única

%\ede{Pontos sobre método de construção: 1) Método a montante é facil de descriminar, }

\caption{Examples of dams of different construction methods within the same region. Figure (a) shows a dam constructed using the Upstream method, while in a nearby region we have the image (b) of a dam constructed using the Single Step method.}
\label{fig:construction_misses}
\end{figure}

Looking at Table \ref{tab:construction_result}, it is possible to notice that all trained models achieved unsatisfactory results in the construction method classification task. Also, unlike other tasks, the results have high confidence intervals. These results can be explained by some factors, such as (1) low spatial resolution, since to distinguish the construction method,it would be necessary to obtain a spatial detailing of the dam contours; (2) the absence of height information on the sensor, the height data in the image is crucial to distinguish different declines of dam heights, and consequently its construction method; (3) dams with different construction methods in the same region, it is very common to have different tailings dams side by side but with different construction methods, as we can see in the examples in figure \ref{fig:construction_misses}.

%\ede{Falar que nao ficou tao boa, intervalo de confiança alta, e provavelmente é dificil porque: resolução baixa para barragens pequenas, barragens do lado do outro}

%é possivel notar que todos os modelos treinados atingiram resultados insatisfatorios na tarefa de classificação de metodos de construção. Além disso, diferente das outras tarefas, os resultados possuem intervalos de confiança altos. Esses resultados podem ser explicados por diversos fatores, como: a baixa resolução espacial, dado que para distinguir o método de construção apenas com a informação visual seria necessario obter um detalhamento espacial dos contornos da barragem; a ausencia da informação de altura no sensor, uma das informações importantes para distinguir diferentes construções de barragem é o declinio dos alteamentos das barragens,

%

These sets of samples end up biasing the final classification model, which receives in its training set almost identical samples with different classes, reducing its effectiveness in the test set.

%\ede{Maneiras diretas de contornar o problema: 1) utilizaçao do sensor LIDAR para informação de elevação (o google earth disponibiliza) 
%2) Filtrar o dataset, diminuido a quantidade de dados porém evitando ambiguidades para o modelo final 
%3) diminuir o tamanho das imagens, concentrando o pouco o contexto e evitando barragens na mesma imagem 
%4) Dados tridimensionais, como nuvem de dados, provavelmente resolveriam essas ambiguidades. }
%

\subsection{Risk Category}

% Nao achei aonde dimiuir uma palavra aqui ainda
The risk of dam failure is measured by considering technical and conservation characteristics. This information is provided by companies to the enforcement agency, as well as emergency action plans and periodic monitoring and conservation reports. In the case of dams that meet Pol\'itica Nacional de Segurança de Barragens (PNSB) criteria, they should also receive periodic visits from enforcement agents. 
However, supervision in Brazil is still limited and heavily dependent on the monitoring of the miners themselves, further increasing the risks of ore mining. \cite{bbc-info}
%\footnote{https://www.bbc.com/portuguese/brasil-47056259}

%Com intuito de contornar esse procedimento de inspeção manual, reformulamos o problema como um problema de classificação de imagens. Nesse caso, dado um subconjunto de imagens de barragens ja rotuladas com suas categorias de risco, tentamos predizer o restante das imagens utilizando apenas a imagem aerea da regiao.

To circumvent this manual inspection procedure, we have reformulated the problem as an image classification problem. In this case, given a subset of dam images already labeled with their risk categories, we try to predict the rest of the images using only the aerial image of the region. 

In this experiment, from a total of 769 images of dams, only 423 were used. This selection was made because of the lack of label of 344 samples and 2 other images were removed to avoid the unbalance problem, since they are the only ones with the label "High". Thus given the selection, there are 423 images of dams divided between the classes: "Low", "Medium". The results obtained in this task can be seen in the Table \ref{tab:risk}.

% \begin{table}[h!]
% \caption{Risk Category}
% \label{tab:risk}
% \tabcolsep=0.155cm
% \begin{tabular}{ccccc}
% \hline
% \textbf{Config}        & \textbf{Mean}   & \textbf{CI}     & \textbf{CI\_MIN} & \textbf{CI\_MAX} \\ \hline
% vgg\_0.0001            & 0.7839          & 0.0448          & 0.7391           & 0.8287           \\ \hline
% inception\_0.0001      & 0.8291          & 0.0380          & 0.7911           & 0.8671           \\ \hline
% \textbf{dense\_0.0001} & \textbf{0.8693} & \textbf{0.0591} & \textbf{0.8102}  & \textbf{0.9284}  \\ \hline
% squeeze\_0.0001        & 0.6626          & 0.0758          & 0.5868           & 0.7383           \\ \hline
% alexnet\_0.0001        & 0.5941          & 0.0424          & 0.5517           & 0.6364           \\ \hline
% resnet\_0.001          & 0.8336          & 0.0436          & 0.7899           & 0.8772           \\ \hline
% \end{tabular}
% \end{table}

% Please add the following required packages to your document preamble:
% \usepackage{booktabs}
\begin{table}[]
\centering
%\caption{Risk Category}

\begin{tabular}{@{}ccc@{}}
\toprule
\textbf{Config}         & \textbf{A.A. (\%)} & \textbf{CI 95\%}       \\ \midrule
VGG 1E-04               & 78.39            & 73.91 - 82.87          \\
Inception 1E-04         & 82.91            & 79.11 - 86.71          \\
\textbf{DenseNet 1E-04} & \textbf{86.93}   & \textbf{81.02 - 92.84} \\
SqueezeNet 1E-04        & 66.26            & 58.68 - 73.83          \\
AlexNet 1E-04           & 59.41            & 55.17 - 63.64          \\
ResNet 1E-03            & 83.36            & 78.99 - 87.72          \\ \bottomrule
\end{tabular}
\caption{This table shows the balanced accuracy of methods on the Risk Category Task.}
%\caption{This table shows the balanced accuracy of methods on the Risk Category Task. The first column contains the methods and learning rate that achieved the best results. The second and third columns show a mean balanced accuracy and 95\% confidence interval obtained from 5-fold cross-validation}
\label{tab:risk}
\end{table}

%In this experiment, we observe high average values, but with large confidence intervals as well. This is a consequence of the general criteria considered for the risk category of a dam. According to Brazilian federal dam safety legislation, risk labels take into consideration: 1) Technical characteristics, 2) Dam conservation status, 3) Safety Plan. While some technical features such as Dam Height, Dam Crown Length are easily visible in terms of images, the Dam Conservation and Safety Plan properties are not used. The absence of this information increases the uncertainty of the model.

In this experiment, it is possible to observe high average values, but with large confidence intervals as well. This is a consequence of the general criteria considered for the risk category of a dam. According to Brazilian federal dam safety legislation, risk labels take into consideration: 1) Technical characteristics, 2) Dam conservation status, 3) Safety Plan. While some technical features such as Dam Height, Dam Crown Length are easily visible in terms of images, the Dam Conservation and Safety Plan properties are not used. The absence of this information increases the uncertainty of the model.

%\ede{Utilização de metadados além da informação do SAR facilmente poderia diminuir o CI}

% Nesse experimento é possível observar valores altos de media, porém com intervalos de confiança grandes também. Isso é uma consequencia dos critérios gerais considerados quanto a categoria de risco de uma barragem. De acordo com a Legislação federal brasileira em segurança de barragens, os rotulos de risk levam em consideração: 1)Características técnicas, 2)Estado de conservação da barragem, 3)Plano de Segurança. Enquanto algumas caracteristicas tecnicas como Altura do barramento, Comprimento do coroamento da barragem são facilmente visiveis em termos de imagens, as propriedades de Conservação da Barragem e Plano de Segurança nao são utilizadas. A ausencia dessas informações torna aumenta a incerteza do modelo.

\subsection{Associated Potential Damage}

%While the risk category of a dam concerns aspects of the dam itself that may influence the likelihood of an accident, its associated potential damage is damage that may occur regardless of its likelihood of occurrence. This damage occurs due to rupture, leakage, soil infiltration or dam malfunction and can be graded according to loss of life and social, economic and environmental impacts.

While the risk category concerns aspects of the dam itself that may influence the likelihood of an accident, its associated potential damage is damage that may occur regardless of its likelihood of occurrence. This damage occurs due to rupture, leakage, soil infiltration or dam malfunction and can be graded according to loss of life, social, economic and environmental impacts.

In this experiment, a total of 425 images of dams were used, since the remains 344 images labels are not available. Images are divided into 3 classes of associated potential damage: Low, Medium, and High. The results obtained can be seen in Table \ref{tab:damage}

% \begin{table}[h!]
% \caption{Associated Potencial Damage}
% \tabcolsep=0.155cm
% \begin{tabular}{ccccc}
% \hline
% \textbf{Config}       & \textbf{Mean}   & \textbf{CI}     & \textbf{CI\_MIN} & \textbf{CI\_MAX} \\ \hline
% vgg\_0.0001           & 0.5463          & 0.0543          & 0.4921           & 0.6006           \\ \hline
% \textbf{inception\_0.0001}     & \textbf{0.6287} & \textbf{0.0343}  & \textbf{0.5943} & \textbf{0.6630} \\ \hline
% dense\_0.001 & 0.5764 & 0.0139 & 0.5625  & 0.5903  \\ \hline
% squeeze\_0.0001       & 0.5092          & 0.0591          & 0.4500           & 0.5683           \\ \hline
% alexnet\_0.0001       & 0.5532          & 0.0455          & 0.5077           & 0.5986           \\ \hline
% resnet\_0.0001        & 0.5596          & 0.0413          & 0.5183           & 0.6009           \\ \hline
% \end{tabular}
% \label{tab:damage}
% \end{table}

% Please add the following required packages to your document preamble:
% \usepackage{booktabs}
\begin{table}[h!]
\centering
\begin{tabular}{@{}ccc@{}}
\toprule
\textbf{Config}         & \textbf{A.A. (\%)} & \textbf{CI 95\%}       \\ \midrule
VGG 1E-04               & 54.63            & 49.21 - 60.06          \\
\textbf{Inception 1E-04}         & \textbf{62.87}            & \textbf{59.43 - 66.30}          \\
DenseNet 1E-03 & 57.64   & 56.25 - 59.03 \\
SqueezeNet 1E-04        & 50.92            & 45.00 - 56.83          \\
AlexNet 1E-04           & 55.32            & 50.77 - 59.86          \\
ResNet 1E-04            & 55.96            & 51.83 - 60.09          \\ \bottomrule
\end{tabular}
\caption{This table shows the balanced accuracy of methods on the Associated Potential Damage Classification Task.}
%\caption{This table shows the balanced accuracy of methods on the Associated Potential Damage Classification Task. The first column contains the methods and learning rate that achieved the best results. The second and third columns show a mean balanced accuracy and 95\% confidence interval obtained from 5-fold cross-validation}
\label{tab:damage}
\end{table}

%\ede{tabelinha, area de contexto da imagem, area indiretamente pelo fluxo do rio}

%Assim como na tarefa de classificação de risco, a complexidade dessa tarefa esta diretamente ligado aos critérios gerais considerados da categoria. Itens como: existência de áreas protegidas, infraestrutura ou serviços, ou equipamentos de serviços públicos essenciais, dificilmente são extraidas em forma de imageamento aereo. Um outro fator muito importante nessa tarefa é a consideração e Existência de população a jusante com potencial de perda de vidas humanas. Como no treinamento da rede neural é levado em consideração apenas um raio de aproximadamente 1.9km (devido ao tamanho da imagem utilizada), todo e qualquer conjunto habitacional fora desse perimetro nao é considerado.

As in the risk classification task, the complexity of this task is directly linked to the general criteria considered in the category. Items such as protected areas, infrastructure or services, or essential utilities equipment are hardly extracted in the form of aerial imagery. Another very important factor in this task is the consideration and existence of downstream population with a potential loss of human life. As the training of the neural network only takes into account a radius of approximately 1.9km (due to the size of the image used), any housing development outside this perimeter is not considered.

\section{Conclusion}
%\todo{The conclusion goes here.}

In this work we introduced BrazilDAM, a novel public dataset based on Sentinel-2 and Landsat-8 satellite images covering all tailings dams cataloged in Brazil. In addition, we provided a benchmark addressing the challenge of land-use and land-cover classification using several modern CNN architectures in the tasks of dam classification, ore classification, constructive method, risk category and associated potential damage. Furthermore, our results suggest that deep neural networks are capable of generalizing well in some proposed tasks (Dam classification, Ore classification, Risk Category), whereas in other tasks (Construction method Associated Potential Damage) it shows poor results, possibly due to the low spatial resolution of images and contextual information considered. To the best of our knowledge, BrazilDAM is the first dataset of Brazil's tailings dams. We hope the dataset encourages the community to develop and test various data-driven algorithms to further boost the state-of-the-arts. As future work, we intend to evaluate the trained models in large scale, over the entire territorial of Brazil.

%We demonstrate how this classification system can be used for detecting land use and how it can assist in improving geographical maps. 

%Experimental results showed that fine tuning tends to be the best strategy in different situations. 

%to other domains, mainly when there are similarities between them (case of UCMerced and RS19 datasets).

%The resulting classification system opens a gate towards a number of Earth observation applications

\section*{ACKNOWLEDGEMENTS}\label{ACKNOWLEDGEMENTS}
The authors would like to thank FAPEMIG, CAPES and CNPq for their financial support to this research project.

{
	\begin{spacing}{1.17}
		\normalsize
		\bibliography{ISPRSguidelines_authors} % Include your own bibliography (*.bib), style is given in isprs.cls
	\end{spacing}
}

\end{document}